\documentclass[10pt, a4paper]{article}
\usepackage{lrec-coling2024} %
\ifdefined\directlua
	\usepackage{fontspec}
\else
	\renewcommand{\rmdefault}{phv}

	\usepackage[letterspace=200]{microtype} 
	
	\usepackage{titlesec}
	 \titlespacing*{\section}
   	 {0pt}{2.ex plus 1ex minus .2ex}{0.8ex plus .2ex}
	\titlespacing*{\subsection}
	{0pt}{2.ex plus 1ex minus .2ex}{1.ex plus .2ex}
\fi
\usepackage{booktabs}
\usepackage{array}
\usepackage{adjustbox} %
\usepackage{makecell} 
\usepackage{pdflscape}
\usepackage{hyperref}
\usepackage{color}
\usepackage{tabularray}
\usepackage{multirow}
\usepackage{natbib}
\usepackage{multibib}
\makeatletter
\def\@mb@citenamelist{cite,citep,citet,citealp,citealt,citepalias,citetalias}
\makeatother
\newcites{languageresource}{~}

\usepackage{graphicx}
\usepackage{tabularx}
\usepackage{soul}
\usepackage{amsmath}
\usepackage{titlesec}
\titleformat{\section}{\normalfont\large\bfseries\center}{\thesection.}{1em}{}
\titleformat{\subsection}{\normalfont\SmallTitleFont\bfseries\raggedright}{\thesubsection.}{1em}{}
\titleformat{\subsubsection}{\normalfont\normalsize\bfseries\raggedright}{\thesubsubsection.}{1em}{}
\renewcommand\thesection{\arabic{section}}
\renewcommand\thesubsection{\thesection.\arabic{subsection}}
\renewcommand\thesubsubsection{\thesubsection.\arabic{subsubsection}}

\usepackage{xcolor}
\usepackage{hyperref}
 \definecolor{darkblue}{rgb}{0, 0, 0.5}
  \hypersetup{colorlinks=true, citecolor=darkblue, linkcolor=darkblue, urlcolor=darkblue}
\usepackage{xstring}
\usepackage{color}
\usepackage{comment}
\usepackage{balance}
\usepackage[hang,flushmargin]{footmisc}
\usepackage{enumitem}
\usepackage{adjustbox}
\usepackage{mdframed}

\definecolor{lightlight-gray}{gray}{0.98}

\global\mdfdefinestyle{rtboxstyle}{%
linecolor=black,%
leftmargin=0cm,rightmargin=0cm,linewidth=0.4pt,
roundcorner=2, skipabove=0.5em, innerleftmargin=5pt, innerrightmargin=5pt,
skipbelow=0pt,backgroundcolor=lightlight-gray
}
\newcommand{\rtbox}[1]{\begin{mdframed}[style=rtboxstyle]{{#1}}\end{mdframed}}

\title{%
    When Do ``More Contexts" Help with Sarcasm Recognition?}

\name{$^\dagger$Ojas Nimase and Sanghyun Hong}
\address{%
    $^\dagger$Westview High School, Oregon State University \\
    $^\dagger$\texttt{ojasnimase@gmail.com, sanghyun.hong@oregonstate.edu}\vspace{1.0em}}

\abstract{
    Sarcasm recognition is %
    challenging because it needs an understanding of the true intention, which is opposite to or different from the literal meaning of the words. Prior work has addressed this challenge by developing a series of methods that provide richer \emph{contexts}, \textit{e.g.}, sentiment or cultural nuances, to models. While %
    shown to be effective individually, no study has systematically evaluated their collective effectiveness. As a result, it remains unclear to what extent additional contexts can improve sarcasm recognition. In this work, we explore the improvements that existing methods bring by incorporating more contexts into a model. To this end, we develop a framework where we can integrate multiple contextual cues and test different approaches. In evaluation with four approaches on three sarcasm recognition benchmarks, we %
    achieve existing state-of-the-art performances and also demonstrate the benefits of sequentially adding more contexts. We also identify inherent drawbacks of using more contexts, highlighting that in the pursuit of even better results, the model may need to adopt societal biases.
}

\begin{document}

\maketitleabstract

\section{Introduction}
\label{sec:intro}

Sarcasm recognition carries importance in various domains, 
ranging from social media analysis~\cite{amir-etal-2016-modelling} 
to product review classification~\cite{parde-nielsen-2018-detecting}.
Beyond its practical applications,
it also offers valuable insights into human behavior. 
For instance, \citet{PERSICKE2013193} use sarcasm recognition 
to investigate the behaviors of individuals on the autism spectrum. 
But recognizing sarcasm is challenging 
because sarcastic expressions %
involve irony, are heavily context-dependent, 
and frequently depend on the tone of speeches~\cite{parde-nielsen-2018-detecting}.
\smallskip

Prior work %
addresses this challenge by integrating \emph{more contexts},
typically sourcing additional information not readily discernible from the training corpus. 
Earlier work~\cite{riloff-etal-2013-sarcasm} proposed learning representations 
(hereafter, we refer to as embeddings) 
that encode the positive or negative meaning of the words 
and use them to %
identify contrasts in a text.
Recent work~\cite{hazarika-etal-2018-cascade} focuses on 
encoding rich contextual information into sentence-level embeddings,
e.g., by combining affective features~\cite{babanejad-etal-2020-affective}
or by leveraging additional training corpus 
to have the embeddings learn contexts implicitly~\cite{10.1007/978-981-15-1059-5_41, liu-etal-2023-prompt}.
\smallskip

While these individual efforts have led to 
significant improvements in sarcasm recognition, 
there is a lack of a systematic study determining 
to what extent each approach is more effective.
It thus remains unclear which methods one should prioritize in using.
It is also unknown what the possibilities and impossibilities are: 
where the failures in sarcasm recognition are attributed 
and if we can address them by developing new methods.
\smallskip

\textbf{Contributions.}
Our contributions are twofold:
\smallskip

\emph{First}, we systematically analyze
the effectiveness of providing additional contexts 
on embeddings in sarcasm recognition.
To run this analysis, 
we design a framework to process
additional contextual information existing work leverages 
when classifying sarcastic texts from non-sarcastic ones.
\smallskip

We apply four different approaches
and evaluate their performances on three sarcasm recognition benchmarks:
IAC-V1, IAC-V2, and Tweets~\cite{oraby-etal-2016-creating, van-hee-etal-2018-semeval}.
Our findings are:
(1) by combining embeddings from the four methods,
we achieve the state-of-the-art performance
shown in the baselines.
(2) sentence-level embeddings are more effective 
than word-level embeddings in sarcasm recognition.
(3) when the embeddings are learned from datasets,
potentially containing more sarcastic texts,
they offer more improvements in recognition.
(4) a training method,
i.e., SimCLR~\cite{chen2020simple},
effective in learning better embeddings
in other domains, %
offer negligible performance improvement.
\smallskip

\emph{Second},
we conduct a manual analysis of the test-set samples,
correctly classified (or incorrectly labeled) by each approach,
and discuss the possibilities and impossibilities of sarcasm recognition.
We observe that the samples are incorrectly classified initially
become correctly classified after we provide more contexts.
We also find the test-set samples 
where we fail to label correctly,
even with all the embeddings combined.
Surprisingly, from our manual analysis,
we show that a model needs to learn societal biases 
to be correct in classifying these samples.
Our result implies that
models may need to learn undesirable biases
or embeddings may require to encode them
to further improve a model's performance 
in sarcasm recognition.

\begin{figure*}[t]
    \centering
    \includegraphics[width=0.9\linewidth]{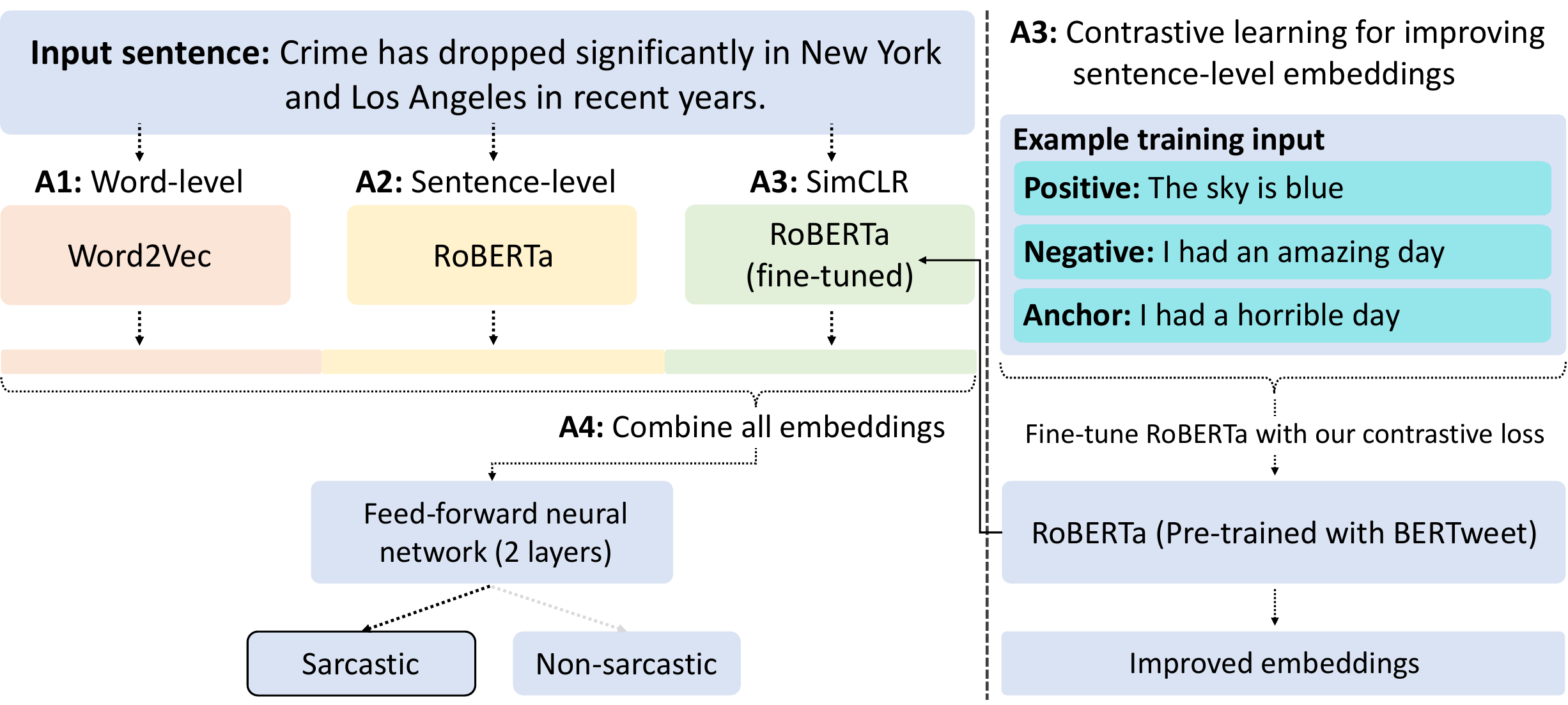}
    \vspace{-0.4em}
    \caption{\textbf{Our framework.} We illustrate how the framework incorporates four different approaches and how we re-train sentence embeddings by adapting a contrastive learning technique~\cite{chen2020simple}. 
    }
    \label{fig:mesh1}
    \vspace{-0.4em}
\end{figure*}

\section{A Framework for Our Study}
\label{sec:method}

Our goal is to study \emph{possibilities and impossibilities} 
when we use more contexts in sarcasm recognition:
how additional contexts have been improving the performance
and what the limits are against pushing the state-of-the-arts.
This section will introduce our method to answer these questions.

\subsection{Methods That Offer More Contexts}
\label{subsec:arch-overview}

We employ (and develop) four methods 
for incorporating additional contexts. 
The first two methods (A1 and A2) 
implement the representative approaches in prior work. 
The next one (A3) studied in different domains, 
but we adopted to sarcasm recognition.
The last method (A4) is the combinations of A1--3,
utilizing their embeddings at once.

\paragraph{A1: Word-level contexts.}
Initial work~\cite{riloff-etal-2013-sarcasm}
leverages word embeddings,
such as Word2Vec or GloVe~\cite{mikolov2013efficient, pennington-etal-2014-glove}
for identifying sarcasm.
We implement this approach in our framework.
Given a sarcastic text, we sum up the embeddings of the words in the text 
and feed them to a classifier to label if the text is sarcastic.
The idea behind this approach is to quantify the contrast between the words.
Positive words are likely to be near another positive word,
and negative words do so; thus, the task becomes identifying 
if negative and positive words are combined together 
to deliver meanings different from literal meanings of words.

\paragraph{A2: Sentence-level contexts.}
The next component of our framework
uses widely-used language models,
based on transformer architectures,
such as RoBERTa~\cite{liu2019roberta}.
These models generate sentence-level embeddings:
they take a sentence (a sequence of words)
and outputs a $k$-dimensional vector.
The typical choice of $k$ is 768.
A standard practice of leveraging these models 
is to pre-train and fine-tune.
We fine-tune a model,
pre-trained on a large corpus of text data,
such as BookCorpus~\cite{7410368}, %
on sarcasm recognition data.
The intuition behind this is:
even if the text, used to pre-train a model,
is from domains different from sarcasm recognition,
it may offer additional contexts to improve the performance.

\paragraph{A3: Improve sentence-level embeddings using contrastive learning.}
We train language models to learn sentence-level embeddings by maximizing (1) agreement between a non-sarcastic text and another unrelated non-sarcastic text, 
and (2) disagreement between the non-sarcastic text and its sarcastic translation.
To learn such embeddings
we adapt a popular contrastive learning framework
(SimCLR), presented by~\citet{chen2020simple}.
Our loss function is formulated as follows:
{\small
\begin{align*}
    \mathcal{L}_{i, j, k} = -\text{log} \frac{\text{exp}(sim(z_i, z_j)/\tau)} {\text{exp}(sim(z_i, z_j)/\tau) + \text{exp}(sim(z_i, z_k)/\tau)}
\end{align*}}
where 
$sim(\cdot)$ is the cosine similarity,
$z_i$, $z_j$, and $z_k$ are the anchor, positive and negative embeddings, and
$\tau$ is a temperature parameter.
In our context,
$z_i$ is the non-sarcastic text, 
$z_j$ is an unrelated non-sarcastic text, 
and $z_k$ is a direct sarcastic translation text of the anchor non-sarcastic text.
Re-training with the loss
allows a model to encode the contexts
that make non-sarcastic and sarcastic sentences 
different in the embedding space.

\begin{table*}[h]
\begin{adjustbox}{width=\textwidth, center}
\begin{tabular}{@{}l|llll|llll|llll@{}}
\toprule
\textbf{Methods} & \multicolumn{4}{c|}{\textbf{IAC-V1}} & \multicolumn{4}{c|}{\textbf{IAC-V2}} & \multicolumn{4}{c}{\textbf{Tweets}} \\ \cmidrule(l){2-13} 
 & \textbf{Acc.} & \textbf{F1} & \textbf{Prec.} & \textbf{Rec.} & \textbf{Acc.} & \textbf{F1} & \textbf{Prec.} & \textbf{Rec.} & \textbf{Acc.} & \textbf{F1} & \textbf{Prec.} & \textbf{Rec.} \\ \midrule \midrule
A1: Word2Vec & 49.8 & 66.5 & 49.8 & 100. & 54.9 & 68.5 & 52.7 & 98.1 & 39.7 & 56.8 & 39.7 & 100. \\ \midrule
A2: RoBERTa & 63.3 & 65.6 & 61.5 & 70.4 & 74.3 & 75.3 & 72.5 & 78.4 & 56.6 & 59.9 & 47.3 & 81.7 \\ \midrule
A2: BERTweet (RoBERTa) & 54.9 & 37.4 & 60.6 & 27.0 & 75.0 & 75.1 & 74.4 & 74.8 & 63.8 & 64.0 & 52.8 & 81.0 \\ \midrule
A3: BERTweet (SimCLR) & 58.3 & 47.0 & 64.1 & 37.1 & 75.2 & 75.2 & 75.2 & 75.1 & 62.8 & 63.4 & 52.0 & 81.4 \\ \midrule
A4: All Embeddings & 72.1 & 72.1 & 71.9 & 72.3 & \textbf{84.0} & 83.2 & \textbf{85.8} & 80.8 & \textbf{82.0} & \textbf{80.2} & 71.3 & \textbf{91.6} \\ \midrule \midrule
\multicolumn{13}{c}{Baselines~\cite{liu-etal-2022-dual}} \\ \midrule \midrule
ADGCN-RoBERTa & \textbf{72.4} & \textbf{72.4} & {72.5} & \textbf{72.4} & 82.1 & 82.1 & 82.2 & 82.1 & 72.2 & 71.4 & 71.3 & 71.9 \\ \midrule
DC-Net-RoBERTa & 69.3 & 69.1 & 69.7 & 69.3 & 83.7 & \textbf{83.7} & 83.7 & \textbf{83.7} & 70.9 & 68.7 & 69.7 & 68.3 \\ \midrule
RoBERTa & 72.1 & 71.9 & \textbf{73.0} & 72.1 & 82.7 & 82.7 & 82.9 & 82.9 & 72.7 & 72.8 & \textbf{72.8} & 73.9 \\ \bottomrule
\end{tabular}
\end{adjustbox}
\caption{\textbf{Performance comparison} of four different approaches (A1--4) to encoding more contexts in sarcasm recognition. We compare accuracy, F1-score, precision, and recall. The bottom two rows are the performance from the baseline approach by~\citet{liu-etal-2022-dual}. Best results are highlighted in \textbf{bold}.}
\vspace{-0.4em}
\label{tab:performance}
\end{table*}

\paragraph{A4: Combine word- and sentence-level embeddings.}
We further combine the embeddings from the above approaches
to leverage full contexts.

\subsection{Putting All Together}
\label{subsec:combining-all}

We finally present a framework 
that enables us to individually (and also comprehensively)
evaluate the effectiveness of our approaches.
The architecture of our framework is shown in Figure~\ref{fig:mesh1}.

\section{Evaluation}
\label{sec:evaluation}

We now empirically evaluate the effectiveness of four different approaches (A1--4). We also analyze samples where each approach can improve upon.

\subsection{Experimental Setup}
\label{subsec:setup}

\paragraph{Datasets.} 
We run our experiments with three benchmarks:
IAC-V1, IAC-V2, and Tweets~\cite{oraby-etal-2016-creating, van-hee-etal-2018-usually}, widely used in sarcasm recognition. Each sentence in the dataset is annotated as sarcasm or non-sarcasm, and we use them as labels. %
We additionally use SarcasmSIGN~\cite{peled-reichart-2017-sarcasm}, composed of sarcastic texts and their multiple, direct, non-sarcastic translations, for contrastive training of sentence embedding models. Note that SarcasmSIGN contains duplicates of non-sarcastic translations, and we filter them out before use.

\paragraph{Models.}
We harness the word embeddings produced by Word2Vec~\cite{mikolov2013efficient}.
To obtain sentence embeddings,
we utilize pre-trained models available from Huggingface. 
Specifically,
we use the RoBERTa-based models~\cite{liu2019roberta}: 
roberta-base\footnote{\href{https://huggingface.co/roberta-base}{https://huggingface.co/roberta-base}} and
vinai/bertweet-base\footnote{\href{https://huggingface.co/vinai/bertweet-base}{https://huggingface.co/vinai/bertweet-base}}.
BERTweet models~\cite{bertweet} undergo pre-training on English Tweets, 
enabling them to learn embeddings from more sarcastic texts.

\paragraph{Metrics.}
We measure the performance using the following four metrics:
classification accuracy (or \emph{accuracy}), 
\emph{F1-score}, \emph{precision}, and \emph{recall}.
\smallskip

Our detailed experimental setup is in Appendix~\ref{appendix:setup-in-detail}.

\begin{table*}[ht]
\begin{adjustbox}{width=\textwidth, center}
\begin{tabular}{@{}l|p{12cm}|l|l@{}}
\toprule
\textbf{Methods} & \multicolumn{1}{c|}{\textbf{Example Texts}} & \textbf{Pred.} & \textbf{Truth} \\ \midrule \midrule
\multirow{2}{*}{A1: Word} & \begin{tabular}[c]{@{}l@{}} I thought God forbid them to eat dead cows, or was it poultry?\\ \ul{emoticonXRolleyes}\end{tabular} %
& S. & S. \\ \cmidrule{2-4} 
 & \begin{tabular}[c]{@{}l@{}} As a gun owner I'm also a property owner.\\ Or are you denying that \ul{guns} are \ul{property}?\end{tabular} %
 & NS. & NS. \\ \midrule \midrule
\multirow{2}{*}{A2: RoBERTa} & See, a \ul{terrorist attack} is probably the sort of thing I would use as an excuse to \ul{not go to work}...%
& S. & S. \\ \cmidrule{2-4}
 & \begin{tabular}[c]{@{}l@{}}So why are you so \ul{afraid of it}?\\ If it is bad you will have a choice to go to a \ul{private insurance} company.\end{tabular} & NS. & NS. \\ \midrule %
\multirow{2}{*}{A2: BERTweet (RoBERTa)} & So everything that other people say on a website or in a book is just and opinion? And everything you say is a fact? \ul{Nice} how you've got \ul{things set up}. & S. & S. \\ \cmidrule{2-4} 
 & Please provide the \ul{actual estimates of the time} required along with the necessary references if you please! \ul{How much time WOULD it take} with confidence limits please! & NS. & NS. \\ \midrule \midrule
\multirow{2}{*}{A3: BERTweet (SimCLR)} & This is just \ul{plain dumb}. Abortion is \ul{NOT the primary means} of birth control. If it is used as birth control, it's \ul{because others have failed or haven't been tried}. & NS. & NS. \\ \cmidrule{2-4} 
 & It's a lot easier to \ul{kill someone} with a \ul{gun than a cigarette or a beer}. & S. & S. \\ \midrule \midrule
\multirow{2}{*}{A4: All Embeddings} & Bravo, Penfold! You are the \ul{neatest} \ul{pricker} of balloons with the \ul{shortest} of \ul{needles} whom I have come across! & S. & S. \\ \cmidrule{2-4} 
 & The idea of \ul{abortion} as population control is \ul{absurd}, especially forced abortions as someone mentioned a few posts ago. Anyone who has read a biology book knows the world has methods of population control on its own, we \ul{don't need to} be doing stuff like that ourselves.
& NS. & NS. \\ \midrule \midrule
\multirow{2}{*}{\begin{tabular}[c]{@{}l@{}}A4: All Embeddings\\(Incorrect predictions)\end{tabular}} & {The tactics \ul{pro-lifers} use make the \ul{Nazis} look like the little league. I mean, seriously. The reason we are dealing with terrorism is because women have the right to the abortion procedure. Wow. Please give me one way those two things relate to each other.} & S. & NS. \\ \cmidrule{2-4} 
 & {The VPC has a political agenda. The \ul{FBI}? That is like saying I believe Coke taste better than Pepsi because the Coke commercial says so.} & NS. & S. \\ \bottomrule
\end{tabular}
\end{adjustbox}

\caption{\textbf{Qualitative analysis.} Each successive method (except for the first and last methods) correctly classifies the samples incorrectly classified by the previous method, we \ul{underline} the parts that seem to cause this performance improvement. S. signifies sarcastic text and NS. signifies non-sarcastic text.}

\vspace{-0.4em}
\label{tab:examples}
\end{table*}

\subsection{Quantitative Evaluation}
\label{subsec:quant-eval}

Table~\ref{tab:performance} summarizes our results.
Overall, we find that using more contexts 
indeed helps with improving the performance in sarcasm recognition.
\smallskip

We first observe that,
when all the embeddings are combined,
we achieve the best performance.
In IAC-V1/-V2, the performances are comparable,
or it is better than the baselines~\cite{liu-etal-2022-dual} in Tweets.
It is an interesting result because 
we achieve performances comparable to the baselines, 
designed explicitly for better sarcasm recognition, 
by simply combining more contexts.
The results suggest that 
the improvements from the prior work 
may come from using more data, 
not from a delicate design of their methodology.
\smallskip

Now we turn our attention to
how much performance improvement each approach brings.
From the first row (A1), 
we see that word-level embeddings 
that encode the contexts from nearby words 
are not sufficient to perform well in sarcasm recognition.
Sentence-level embeddings (A2)
that capture contexts from long-range dependency 
significantly improve performance.
The performance further increases
when models are pre-trained on a corpus (A3),
potentially including more sarcastic texts, 
such as English Tweets.
Contrastive learning,
in \emph{contrast} to the advances made in other domains,
does not improve the performance more.
Instead, if we use all the embeddings,
this straightforward approach leads us to the best (A4).

\subsection{More Does Not Always Mean Better}
\label{subsec:qualitative-eval}

We now manually analyze the samples correctly classified by an approach but misclassified by the preceding one.
Previous studies have relied on amortized metrics, e.g., accuracy, to quantify performance improvements. 
While shown effective, they often leave ambiguity regarding whether the claimed improvements result from the proposed techniques or if other factors contribute to the increased accuracy.
Our manual, per-sample analysis is a starting point to take an in-depth look at existing approaches and whether they are improving as shown in their original studies.
We find in our qualitative analysis that in some cases, the improvements come as claimed, but in other cases, it is from undesirable model behaviors like biases.
Examples are shown in Table~\ref{tab:examples}.

\paragraph{A1 and A2.} 
We show how sentence embeddings (A2) enhance 
a model's understanding of sarcastic texts.
For example, the second two rows show that 
an approach solely relying on word embeddings 
cannot capture the long-range dependency between, 
e.g., ``terrorist attack'' and ``not go to work.'' 
A1 classifies the example text as non-sarcasm.
If sentence embeddings learn from texts
potentially containing sarcasm (A2: BERTweet), 
the embeddings of ``And everything you say is a fact?" 
are not similar to those of genuine questions. 
They are closer to accusatory questions.

\paragraph{A3.}
If the models that generate sentence embeddings 
are further fine-tuned using the contrastive learning approach, we observe that the embeddings begin to encode paraphrases
commonly found in sarcastic texts, such as 'plain dumb.' 
But, in terms of performance, these advancements result in only a marginal difference (see Table 1~\ref{tab:performance}).

\paragraph{A4.}
If we combine all the embeddings, 
we see the advantage of using information
from both word-level and sentence-level embeddings. 
In word embeddings, the ``neatest picker'' and ``shortest of needles" 
contain two words with an opposite sentiment.
However, just looking at individual phrases 
is not sufficient to identify a sarcastic tone, 
and the sentence-level embeddings enable 
connecting the two and recognizing the sarcasm.

\paragraph{Biases.}
We further analyze the failure cases of A4 
and find that, to make these sample texts correctly classified, 
embeddings may need to encode undesirable biases.
For example, in the last two rows, 
to understand the sarcasm in the first text,
a model may need to have negativity toward ``pro-lifers'' to align it with ``Nazi.''
The same goes for the second example.
A model (or embeddings) may need to be biased against conspiracy theorists 
to correctly classify the sample text as sarcasm.
\smallskip

More analyses can be found in Appendix~\ref{appendix:more-analysis}.

\section{Conclusion}
\label{sec:conclusion}

This paper studies the role of rich contextual information in sarcasm recognition. 
To conduct this study, we develop a framework that implements four representative approaches to incorporating richer contexts for sarcasm recognition. 
By evaluating these approaches on three sarcasm recognition benchmarks, we provide a new viewpoint on long-held beliefs in sarcasm detection. 
We show that: (1) Just combining more embeddings will offer the same performance in sarcasm detection as using complex model architectures or delicate training methods. 
(2) Pushing the performance further may require a model to learn undesirable biases, necessitating rethinking whether we should keep improving the current approaches.
\smallskip

\textbf{What's Next?}
Our work underscores the need for future work 
to develop new methodologies for building models 
that excel in sarcasm detection 
and minimize reliance on undesirable biases, 
such as those related to gender or societal norms.
To achieve this goal, 
we encourage the following directions for future research:
(1) A systematic investigation of when and how these biases are introduced to a model. 
This involves adapting existing metrics or devising new ones to accurately quantify biases present in models.
Moreover, we envision developing a novel method to determine which training instances significantly impact the accurate identification of specific sarcastic expressions.
(2) In light of the remarkable capabilities of large-language models,  future research should assess whether increasing model size effectively addresses the bias issues we have identified.
Although our manual analysis suggests that improvements might inadvertently depend on undesirable biases, the efficacy of scaling as a solution remains uncertain.
It is therefore important to empirically test this hypothesis to determine the viability of scaling as a strategy to mitigate bias.
(3) Moreover, future work may focus on the cross-collaboration that must occur between research and social institutions.
Given that enhanced sarcasm detection 
may inadvertently learn and propagate undesirable biases, 
it is important to prevent the deployment of 
such biased models within social institutions.
Moreover, considering that much of the training data 
for sarcasm detection models comes from 
social media (e.g., Twitter), 
there is a need for researchers to collaborate with 
these companies to limit the introduction of harmful data.
By working together, we hope to develop 
ethical and unbiased 
natural language processing methods.

\section{Acknowledgements}

We thank anonymous reviewers for their valuable feedback.
This work is partially supported by 
the Samsung Global Research Outreach (GRO) Program 
and the Google Faculty Research Award.

\nocite{*}
\section{Bibliographical References}
\label{sec:reference}

{
    \bibliographystyle{lrec-coling2024-natbib}
    \bibliography{bib/thiswork}
}

\appendix

\section{Experimental Setup in Detail}
\label{appendix:setup-in-detail}

Here we describe our experimental setup in detail
for the reproducibility of our analysis results.
Our code is available at \href{https://github.com/Secure-AI-Systems-Group/sarcasm-detection}{https://github.com/secure-ai-systems-group/sarcasm-detection}.

\paragraph{A1: Word-level embeddings.}
We employ the Word2Vec model from Gensim~\cite{rehurek_lrec} to generate text embeddings. We first convert each word within a text into its corresponding embeddings and concatenate them. We limit each text sample to a maximum of 50 words.

\paragraph{A2: Sentence-level embeddings}
are generated by using the \verb|vinai/bertweet-base| and \verb|roberta-base| models. %
We use the AutoTokenizer class for our models. We feed the token-level embeddings, generated by the tokenizer to our models. We average the embeddings obtained from our model to make a single sentence embedding. %

\paragraph{A3: Contrastive-learning process.}
To further improve the quality of sentence-level embeddings, we fine-tune the vinai/bertweet-base model via our contrastive learning technique:

\begin{enumerate}[label=(\arabic*), noitemsep, topsep=0.3em]
    \item We first fine-tune the \verb|vinai/bertweet-base| model on the SarcasmSIGN dataset~\cite{peled-reichart-2017-sarcasm}. Before fine-tuning, we remove duplicates from the dataset.
    \item We adapt the contrastive learning framework presented by~\citet{chen2020simple} for visual representations to enhance our sentence-level embeddings. The framework uses a 2-layer feedforward neural network to produce 256-dimensional representations; we follow this process and decrease the representation dimension from 768 to 256. We then fine-tune these two models using \textit{NT-Xent} loss.
    \item We fine-tune the model for 10 epochs, using a temperature of 0.7, batch size of 50, a learning rate of 1e-5, a weight decay of 1e-3, and the AdamW optimizer. 
    \item We follow the same process outlined in A2 with this fine-tuned \verb|vinai/bertweet-base| to create the sentence-level embeddings.
\end{enumerate}

\paragraph{Sarcasm recognition models.} 
We employ a 2-layer feedforward neural network to implement our models. When we test each approach individually, we set the dimension of the input linear layer to 768. We set the hidden layer dimension to 128 and employ the ReLU activation function for non-linearity. The final linear layer classifies the text as sarcastic or non-sarcastic. We train them for 5 epochs using a weight decay of 0.01, a learning rate of 1e-5, and a batch size of 32. We use the cross-entropy loss and AdamW optimizer. 
\smallskip

When all four types of embeddings are used,
we concatenate them, and therefore,
the input layer's dimension increases from 768 to 39936. The A1, A2 RoBERTa, and A3 embeddings have already been generated and are fed in as lists while A2 BERtweet embeddings are added by incorporating 'vinai/bertweet-base' with the model. A linear layer is used to reduce this dimensionality back to 768 and the other architectural choices are kept the same.
We train this model for 5 epochs with a batch size of 16, a weight decay of 0.01, and a learning rate of 1e-5.
We employ the F-$\beta$ score as our loss function.
We use the AdamW optimizer.

\section{More Qualitative Analysis}
\label{appendix:more-analysis}

Here we provide more examples of a model learning biases for improving sarcasm recognition.

\rtbox{
\begin{enumerate}[label=(\arabic*), itemsep=0.1em, topsep=0.3em] 
    \item Katie pisses me off so bad \#TheApprentice
    \item @cnsnews Obama and Hillary convinced Ukraine that they would protect them if they essentially disarm. Need to keep at least one promise.
    \item Everytime I try to like Chris Brown he does something to royally eff that up. Dude is a chronic loose cannon \#chrisbrown \#Karrueche
    \item Again, as an ignorant layman, I can only get the gist of this material, but how anyone could possibly argue against the genetic code as a product of intelligent design is beyond me.
\end{enumerate}
}
\noindent\textbf{Examples correctly classified by a model potentially leaned societal biases.} We showcase four examples, incorrectly classified by the models in A1--3, while correctly classified by our A4 model.
\vspace{1.em}

We showcase example texts (1)--(3) that were incorrectly classified as sarcastic by A1--A3 and correctly classified by A4 as non-sarcastic. (4) was incorrectly classified as non-sarcastic by A1-A3 and correctly classified as sarcastic by A4. We conduct a manual analysis on why: 
(1) shows A4 may become biased against Katie, a contestant in ``\#TheApprentice". 
(2) shows A4 may become biased against ``Obama and Hilary" to correctly classify it.
(3) %
shows A4 may become biased against ``Chris Brown".
(4) %
shows A4 may become biased against ``Intelligent Design" and mock it.

\end{document}